\documentclass{article}

\usepackage[preprint]{neurips_2023}

\usepackage[utf8]{inputenc} 
\usepackage[T1]{fontenc}    
\usepackage{hyperref}       
\usepackage{url}            
\usepackage{booktabs}       
\usepackage{amsfonts}       
\usepackage{nicefrac}       
\usepackage{microtype}      
\usepackage{xcolor}         
\usepackage{graphicx}
\usepackage{amsmath}

\title{Training Dynamics of Contextual N-Grams in Language Models}

\author{
  Lucia Quirke$^*$\\
  Independent \\
  \texttt{luciarosequirke@gmail.com} \\
  \And
  Lovis Heindrich$^*$\\
  Independent \\
  \texttt{heindrich.lovis@gmail.com} \\
  \AND
  Wes Gurnee \\
  MIT \\
  \And
  Neel Nanda \\
  Independent \\
}

\begin{document}

\maketitle
\def\thefootnote{*}\footnotetext{These authors contributed equally to this work. Order was randomized.}
\def\thefootnote{\arabic{footnote}}

\begin{abstract}
Prior work has shown the existence of contextual neurons in language models, including a neuron that activates on German text. We show that this neuron exists within a broader \textit{contextual n-gram} circuit: we find late layer neurons which recognize and continue n-grams common in German text, but which only activate if the German neuron is active. We investigate the formation of this circuit throughout training and find that it is an example of what we call a second-order circuit. In particular, both the constituent n-gram circuits and the German detection circuit which culminates in the German neuron form with independent functions early in training---the German detection circuit partially through modeling German unigram statistics, and the n-grams by boosting appropriate completions. Only after both circuits have already formed do they fit together into a second-order circuit. Contrary to the hypotheses presented in prior work, we find that the contextual n-gram circuit forms gradually rather than in a sudden phase transition. We further present a range of anomalous observations such as a simultaneous phase transition in many tasks coinciding with the learning rate warm-up, and evidence that many context neurons form simultaneously early in training but are later unlearned.
\end{abstract}

\section{Introduction}
Mechanistic interpretability is a growing field that studies internal mechanisms of neural networks \citep{olah2022mechanistic, räuker2023transparent}. These internal mechanisms are often represented as circuits, composed of features---specific properties of the input---connected by weights \citep{cammarata2020thread:}. Current work in mechanistic interpretability tends to focus on understanding circuits in fully trained language models \citep{wang2022interpretability, gurnee2023finding}. However, neural networks are formed via the continuous process of gradient descent, so a natural line of inquiry for better understanding trained models is understanding the path they took to get there. Many fascinating aspects of model behavior are inherently tied to training dynamics, such as the phenomenon of phase transitions and emergence, where a quantitative change to a model results in a qualitative change in the model's capabilities \citep{wei2022emergent}. One hypothesis is that emergent capabilities correspond to specific circuits forming suddenly throughout training \citep{michaud2023quantization}, and a natural question is whether by investigating the dynamics of circuit formation we can better understand emergent capabilities.

\begin{figure}
    \centering
    \includegraphics[width=0.95\linewidth]{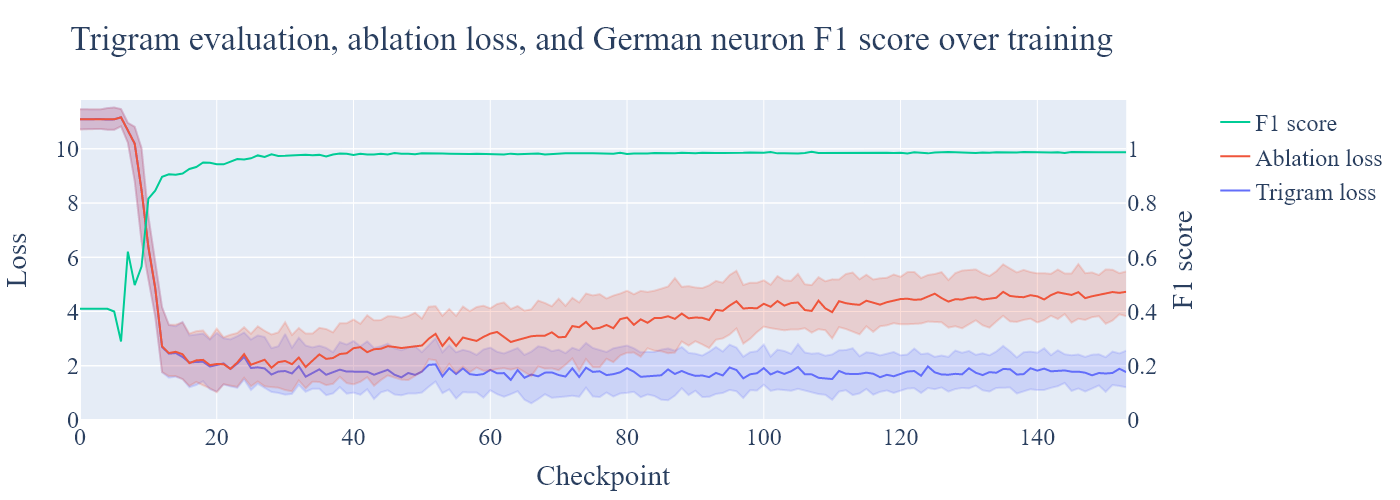}
    \caption{Key circuit metrics over training. The trigram loss represents the model's average performance on predicting our dataset of German trigrams, and the ablation loss demonstrates the model's increasingly degraded performance on the same data when the German neuron is ablated. Shaded areas represent the 25th to 75th percentile confidence intervals. The F1 score shows that the German neuron develops before the trigrams' dependence on it.}
    \label{fig:full_result}
\end{figure}

In this work we studied circuits involving transformer model neurons. Transformer neurons can be classified as either \textit{monosemantic}, responding to a single feature, or \textit{polysemantic}, responding to multiple \citep{toy_models}. While the typical neuron in a transformer is polysemantic \citep{bills2023language, elhage2022solu}, monosemantic neurons have been documented in both transformers and other neural network architectures \citep{toy_models, burns2022discovering, cammarata2020thread:, elhage2022solu}. \citet{gurnee2023finding} discovered a family of monosemantic neurons in transformer models which represent high-level contextual features such as the input text language. Strikingly, they found that these individual language neurons play a crucial role for predicting text in the represented language, leading to a large decrease in performance when ablating the neuron. 

We identified circuits that depend on one such neuron in the language model Pythia 70M \citep{pythia} that corresponds to German input text. One natural guess for the role of a German context neuron is to boost German unigrams, that is, increasing the output logits for tokens that are more frequent in German text. We identified this behavior, but we also discovered a large indirect effect \citep{mcgrath2023hydra}, where later model components compose with the context neuron to form circuits. We studied one family of these circuits, which we call \textbf{contextual n-grams}: late layer neurons that detect and continue n-grams common in German, but only if the German context neuron in a prior layer has activated.

To understand how the model learns contextual n-grams, we investigated properties of the circuit over the course of training using the many checkpoints available for the Pythia models. We discovered that the model first learns to represent two simple circuits which independently contribute to model performance, the German n-gram circuit and a German detection circuit culminating in the context neuron, in a phase transition early in training. Then, later on in the training process, the model slowly connects the two circuits to form the contextual n-gram circuit such that the n-gram depends on the activation of the context neuron. Notably, the contextual n-gram circuit itself does not seem to form in a phase transition, and does not increase n-gram prediction performance beyond what was learned in the n-gram phase transition. This presents a mystery around why the circuit forms at all. 

Although this is not a primary focus of our paper, we speculate that the contextual n-gram circuit is useful to the model to reduce interference from superposition. N-grams are a plausible candidate to have high levels of superposition because of their abundance and sparse activation patterns \citep{gurnee2023finding}. This can result in interference where unrelated text may incorrectly partially activate the n-gram, adding noise to the model. Ensuring that the n-gram can only activate in the presence of German text reduces incorrect activations.

\section{Methods}

\paragraph{Model \& Dataset} We restrict our investigation to one of the smallest models (70M parameters) in the Pythia series of generative pre-trained (GPT) language models \citep{pythia}, as used in \citet{gurnee2023finding}. Our evaluation is run on a subset of European Parliament multilingual data from The Pile \citep{pile}, the public dataset used to train the Pythia series. Specifically, we constructed validation sets of 2,000 examples for German and English examples from the European Parliament subset of The Pile. 

\paragraph{Identifying German Neurons} Following \citet{gurnee2023finding}, we use sparse probing to detect monosemantic German language neurons. For each neuron in the model, we collect a dataset of $10,000$ neuron activation values for both English and German text, resulting in a balanced dataset of $20,000$ examples. We randomly select a training set of $16,000$ examples from the dataset and then train a binary 1-dimensional logistic regression probe which maps the neuron's activation to the "is German" label using a learned threshold. This method does not require hyperparameters. We evaluate the probe on the remaining $4,000$ test examples using an F1 score \citep{taha2015metrics}. \footnote{We also calculate Matthew's correlation coefficient \citep{why_mcc} with similar results.} 

\paragraph{Estimating German Neuron Importance} To test the importance of the German context neurons for predicting German text, we make use of mean ablation \citep{wang2022interpretability, vig2020investigating}, a method in which individual neurons are ablated by setting their activation value to their average activation during the forward pass. This has the effect of removing information which varies in a reference distribution, while preserving constant biases that other components may rely on. Specifically, we pin the German context neurons' activations to their mean activations on English text. We then compute the increase in cross-entropy loss when ablating one or more German context neurons over the loss of the original model. 

\paragraph{Dataset examples} To find contextual n-grams, we search for dataset examples where ablating the German context neuron leads to a large loss increase through indirect effects. As this measure alone does not necessarily result in interesting n-grams, we additionally filter for trigrams with a high overall loss increase, a low loss when the German context neuron is not ablated, and an indirect loss increase that is larger than the direct loss increase. We verify that our discovered tokens are in fact trigrams by generating random datasets to test and filter them in isolation. Random prompts are generated by sampling $20$ tokens from the $100$ most frequent German tokens and then appending the trigram that is being tested. We note that this cherry-picks n-grams where the German neuron matters, and so we do not make a general statement about \textit{all} n-grams.

\paragraph{Direct and indirect effects}
The total effect of the German context neuron on the model's outputs can be decomposed into its \textit{direct} and \textit{indirect} effects \citep{pearldirect2001}. A neuron's direct effect is defined as its modification of token probabilities in the output logits via the residual path, avoiding all subsequent layers. Its indirect effect is the complement of the direct effect, comprising all modifications to the output token probabilities mediated by paths through subsequent feed-forward or attention layers. To compute the direct effect, we first run the model and cache the activations of all layers after the context neuron. Then, we run the model again while mean ablating the context neuron and patching in activations of later components from the previous run. To compute the indirect effect, we follow a similar setup: we first run the model while mean ablating the German context neuron and cache the activations, then we run the model a second time while patching in the cached activations but without ablating the context neuron. This follows the method in \citet{mcgrath2023hydra}.

\paragraph{Direct Logit Attribution} Additionally, we use direct logit attribution (DLA) \citep{elhage2021mathematical, nostalgebraist2020} to understand the direct effect of the context neuron and determine which output tokens in the model's vocabulary are directly boosted or inhibited by the neuron's activation. DLA is calculated by taking the dot product of a neuron's output weight and the unembedding matrix, resulting in one scalar value per output token that describes the neuron's effect on the token. We note that this is computed purely from the weights, and does not depend on the activations on any specific input. We collect a dataset of the 200 most frequent tokens for both German and English text in the European Parliament data, while filtering out tokens that do not contain at least one alphabetic character (i.e. removing both number and punctuation tokens). We additionally exclude tokens frequent in both languages, resulting in 188 frequent tokens for each language. We identify neurons which tend to directly boost German or English tokens by calculating the average logit increase over these tokens.

\section{Results}
Our main results are summarized in Figure~\ref{fig:full_result}. Focusing on the most impactful context neuron (\textit{L3N669}), we found that both the context neuron and the ability to predict many German trigrams were learned early in training in a phase transition around checkpoint 10. However, in earlier checkpoints before checkpoint 30, the trigram loss increase from ablating the context neuron was negligible, indicating that the model first learned the trigram without utilizing the context neuron. Throughout the rest of training, the loss increase gradually rose to over 100\% of the trigrams' steady-state loss, indicating the model learned to depend on the German context neuron only later in training. In the remainder of this section, we present evidence for all three of our claims, then highlight additional interesting phenomena we observed.

\paragraph{Trigrams and context neurons are learned in a simultaneous phase transition}

\begin{figure}
    \centering
    \includegraphics[width=0.95\linewidth]{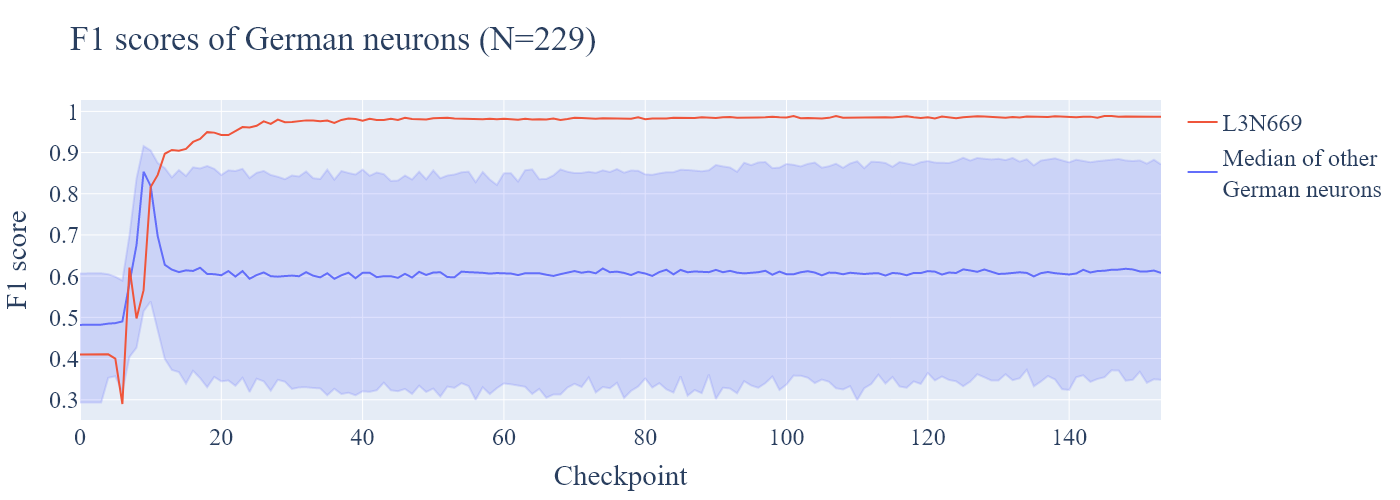}
    \caption{Median context neuron F1 scores for distinguishing between English and German text over training checkpoints. Only neurons with a maximum F1 higher than $0.85$ are shown ($N=229$). The shaded area in the plot shows the 5th to 95th percentile. }
    \label{fig:f1_scores}
\end{figure}

Figure~\ref{fig:full_result} shows the trigram loss curve for the German trigrams on which the context neuron has a high indirect effect across all checkpoints of the Pythia model. We observed that almost all trigrams are learned at the same time, around checkpoint 10. Afterwards, the loss curves plateaued and only improved marginally over the rest of training. This coincided with a general phase transition in which the model acquired most of its general capabilities for both English and German text (see Figure~\ref{fig:phase_transition} in the appendix).


During the same phase transition, we also observed the formation of German context neurons. Using our probing setup, we searched for German context neurons throughout training checkpoints. Figure~\ref{fig:f1_scores} shows the F1 score of the most accurate German language-specific neurons throughout training. We found a surprisingly large number of highly accurate German context neurons ($229$ neurons with a maximum F1 of $0.85$ or higher at some point throughout training). Congruent with our previous finding on the general loss curves, we found that the context neurons were mainly learned during the same phase transition around checkpoint 10. Interestingly, most of the early context neurons were only context neurons for a very short period of time before being unlearned, as can be seen in the steep drop-off in F1 scores between checkpoints 10 and 15 in Figure~\ref{fig:f1_scores}.  

The neuron with the highest F1 score is neuron 669 in layer 3 (\textit{L3N669}). While \textit{L3N669} was also mostly learned during the phase transition, its F1 score did not peak at checkpoint 10 as we observed in most other neurons (see Figure~\ref{fig:f1_scores}). Instead, it continually improved how much its activations separate German and English inputs up to checkpoint 30 (F1 of $0.97$). 

\begin{figure}
    \centering
    \includegraphics[width=0.95\linewidth]{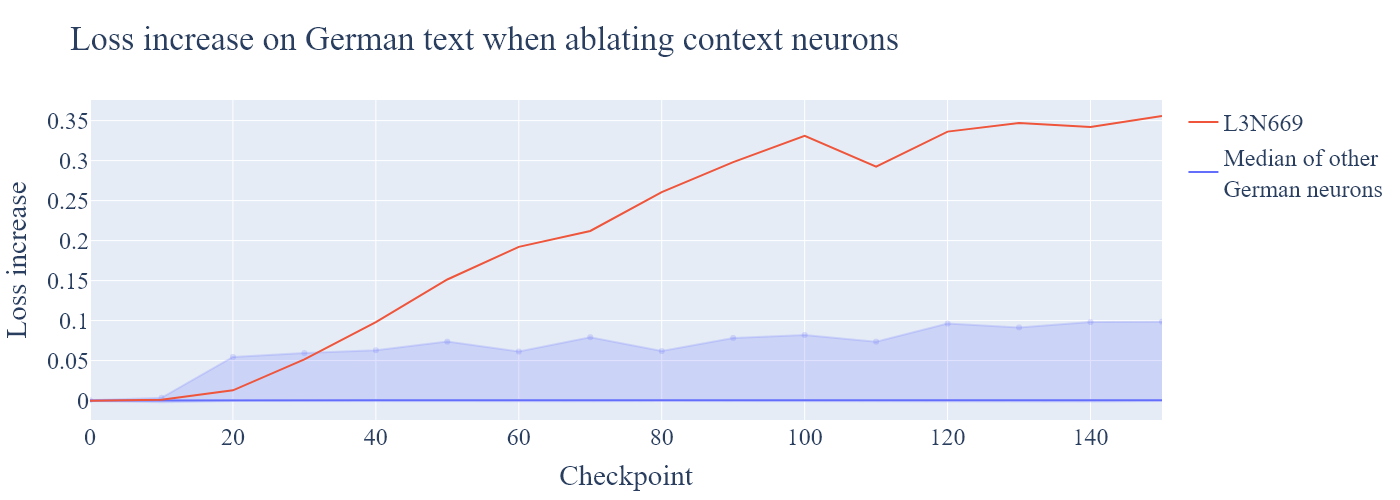}
    \caption{Average increase in loss on German text when ablating any other neuron which achieves an F1 > 0.85 at any point during training (229 in total), compared to L3N669. Loss increases are sampled every 10 checkpoints. The shaded area shows the full range of values.}
    \label{fig:ablation_increases_over_time}
\end{figure}

\paragraph{Context neurons' impact on text prediction}
We determined the importance of the neurons we found by individually mean ablating each neuron over the training checkpoints and measuring the increases in loss. Of the $229$ neurons, most did not have any noticeable effect on the loss for either English or German text prediction, raising the question of why the model first learned such a large number of poorly utilized context neurons. Figure~\ref{fig:ablation_increases_over_time} shows the average loss increase on German text when ablating each of the context neurons throughout checkpoints. The neuron most relevant for predicting German text is \textit{L3N669}, while most others did not lead to a large increase in loss.

Zooming in on the most relevant context neuron (\textit{L3N669}), we observed a large loss increase of over $10\%$ from mean ablating the neuron on German text in the fully trained model. While the neuron is already capable of detecting the language early on in training (it reaches an F1 of $0.9$ by checkpoint 13), its impact as measured by mean ablations is quite low in early checkpoints. The ablation effect increases throughout training, ranging from $0.4\%$ at checkpoint 20 to over $10\%$ at checkpoint 100. This indicates that while the neuron was learned early during training, the model gradually learned to utilize the neuron throughout training.



\paragraph{Contextual trigrams develop gradually}
Figure~\ref{fig:full_result} shows the model's performance on the German trigram dataset as well as the model's reliance on the context neuron when predicting the trigrams. When the model achieved the ability to predict the German trigrams early in training, they did not depend on the German context neuron, as ablating the context neuron did not majorly impact the model's trigram predictions. In contrast, trigram performance in the fully trained model was highly dependent on the context neuron. This highlights an interesting phenomenon happening during model training: two features that are learned early on are later combined in a gradual process. We call this a \textbf{second-order circuit}, a circuit that is made of pre-existing components.

\paragraph{Direct \& indirect effects} 

\begin{figure}[ht]
    \centering
    \includegraphics[width=0.95\linewidth]{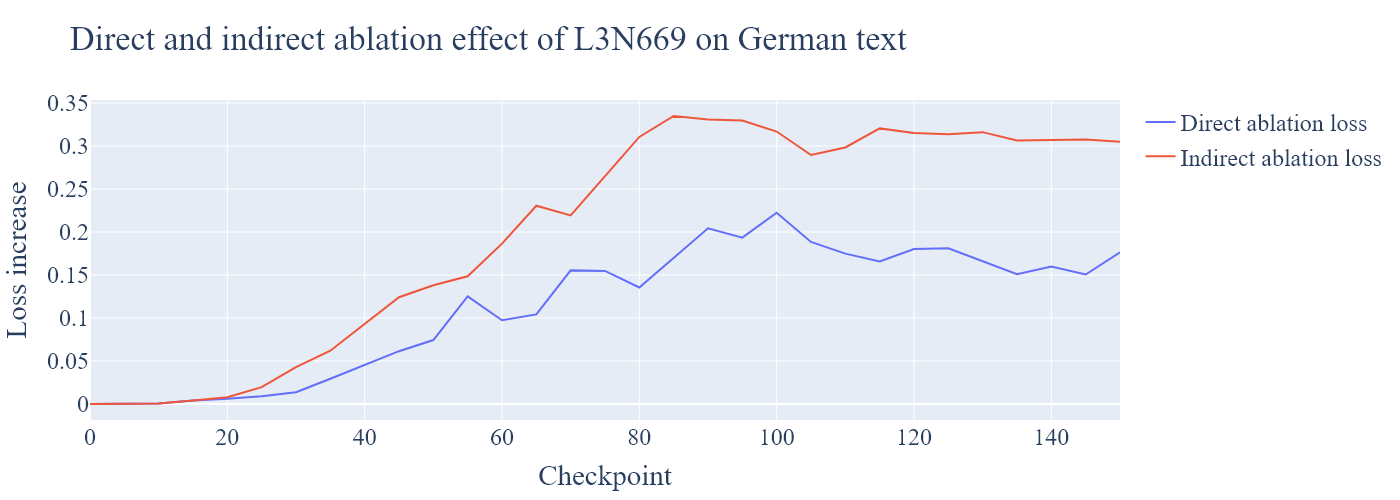}
    \caption{Average direct and indirect effects of ablating L3N669 over checkpoints. Losses are evaluated every 5th checkpoint.}
    \label{fig:indirect_effects}
\end{figure}

\begin{figure}[ht]
    \centering
    \includegraphics[width=0.95\linewidth]{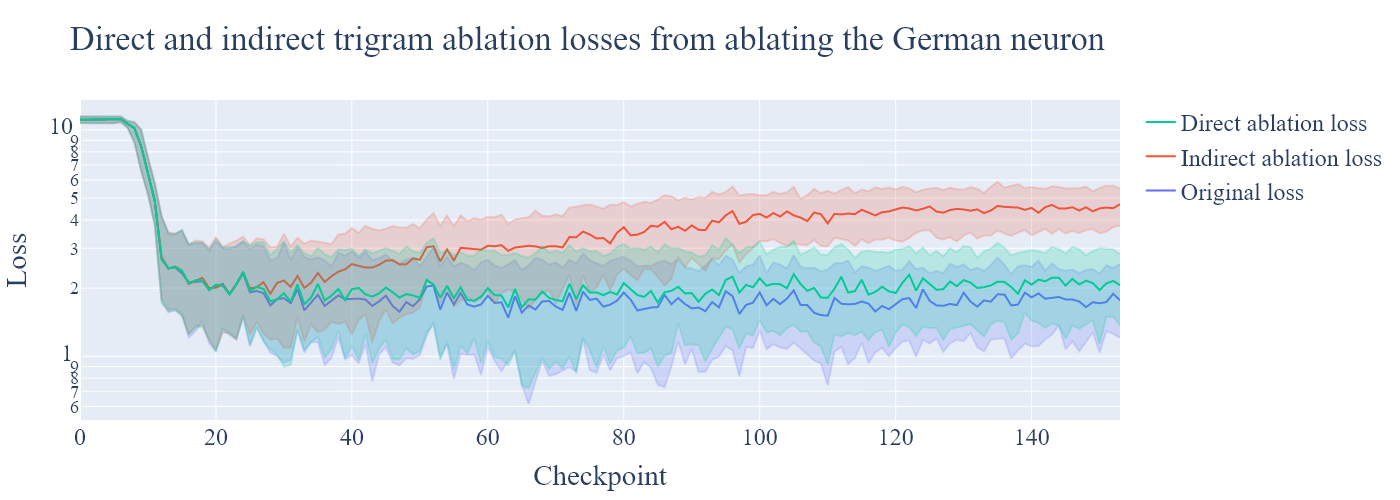}
    \caption{Median loss values for the trigram dataset ($N=189$), with ablation losses for the German context neuron split into direct and indirect effects. The shaded area in the plot shows the 25th to 75th percentile.}
    \label{fig:trigram_losses_direct_indirect}
\end{figure}

\begin{figure}[ht]
    \centering
    \includegraphics[width=0.95\linewidth]{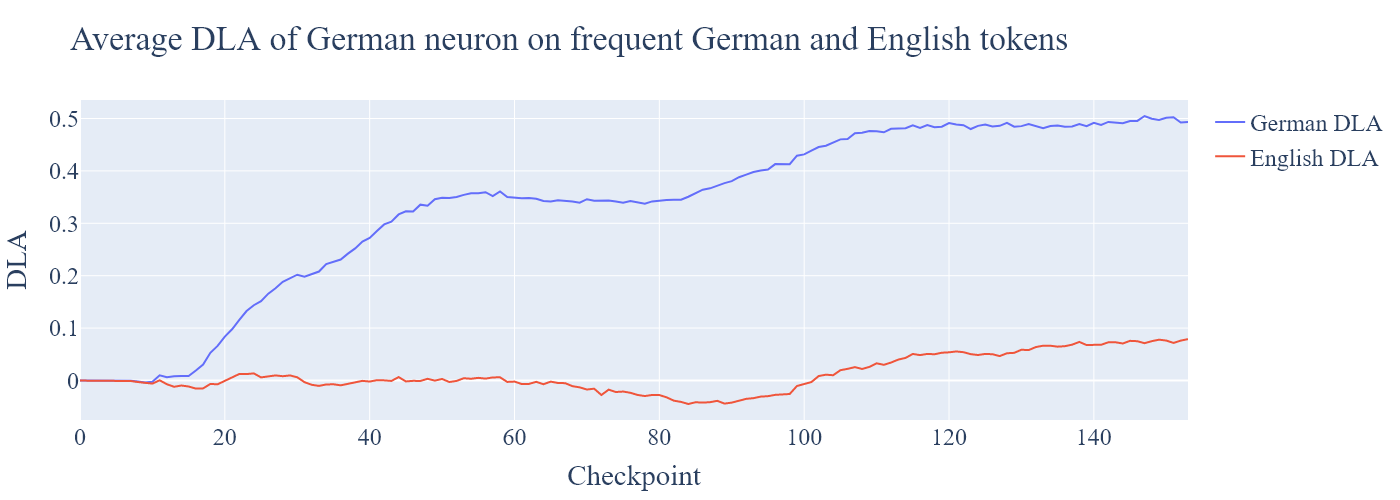}
    \caption{Difference in average direct logit attribution of \textit{L3N669} between frequent German and English tokens.}
    \label{fig:dla}
\end{figure}

We calculated the proportions of direct and indirect effects of neuron \textit{L3N669}, and found large increases in loss coming from both direct and indirect effects, as depicted in Figure \ref{fig:indirect_effects}. We verified that for our trigram dataset, the increase in loss stems from the indirect effect of the German context neuron, as it forms contextual n-gram circuits with later neurons in the model (see Figure~\ref{fig:trigram_losses_direct_indirect}). 

As the context neuron also had a significant impact on loss through its direct effect, we determined the token probabilities directly increased by the neuron by computing its direct logit attribution to frequent German and English tokens (see Figure~\ref{fig:dla}). While the boost applied to frequent English tokens is close to zero throughout training, German tokens are strongly boosted. This indicates that the role of the context neuron is twofold; it boosts frequent German tokens directly, and indirectly enables more complex circuits such as contextual n-grams.

\paragraph{General phase transition}

\begin{figure}[ht]
    \centering
    \includegraphics[width=0.95\linewidth]{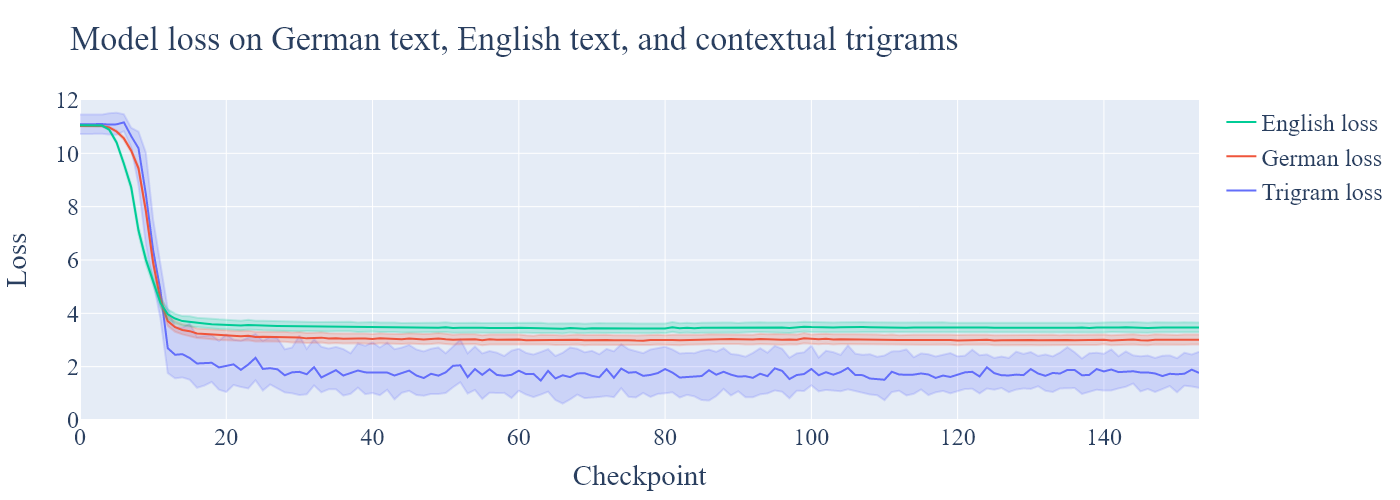}
    \caption{Overall model loss on English and German data. The shaded area in the plot shows the 25th to 75th percentile.}
    \label{fig:phase_transition}
\end{figure}

We found that a large proportion of the model's capabilities emerge rapidly in a general phase transition. Figure~\ref{fig:phase_transition} shows that the model rapidly learned to predict English text as well as German text around checkpoint 10. During the same phase transition, the model also learned to predict the dataset of German trigrams, although the trigrams do not depend on the context neuron at that point. We believe this phase transition might partially be explained by the linear learning rate warm-up used in the Pythia models \citep{pythia}.

\section{Related Work}

\paragraph{Circuits analysis}
Our use of circuits as a unit of analysis was inspired by the circuits-style approach to understanding model properties of \citet{cammarata2020thread:}. 

Causal mediation analysis was first used to analyze neural language models by \citet{vig2020investigating}, and further developed for circuits analysis by \citet{wang2022interpretability} and \citet{mcgrath2023hydra}. Ablations are commonly used to test the causal role of model components in implementing neural network behaviors \citep{chan2022causal, leavitt2020falsifiable}. Our method of ablating a feature using a mean calculated over a restricted subset of the training distribution has commonalities with the mean ablations in \citet{wang2022interpretability} and the pinned activation values in \citet{bricken2023monosemanticity}.

Previous circuit analyses using attention heads as a unit of analysis found attention heads in shallow transformers which model skip-trigram statistics \citep{elhage2021mathematical}, and many interpretable attention heads in larger models including the induction heads \citep{olsson2022incontext}, name mover heads, and s-inhibition heads \citep{wang2022interpretability} used in the indirect object identification circuit. Several results have discovered features corresponding to neurons or directions in the activation spaces of transformer feed-forward layers \citep{meng2023locating, gurnee2023finding}, but work focused on interactions between features in feed-forward layers has been limited to CNNs \citep{cammarata2020thread:}. \citet{geva-etal-2021-transformer, geva2022transformer} proposed that the feed-forward layers of transformers function as key-value memories, where each key corresponds to a feature of the input which when activated makes an additive update to the output distribution, which is consistent with both the contextual neuron and contextual n-grams we have found (though the contextual n-grams are not aligned with the neuron basis). 

\paragraph{Training dynamics}

Several works \citep{olsson2022incontext, nanda2023progress} discovered circuits in language models that seem causally relevant to emergent capabilities and can be ablated to reduce the capabilities. \citet{michaud2023quantization} postulated that many capabilities in language models emerge rapidly in phase transitions and proposed a model in which any capabilities which appear to instead form gradually can be decomposed into discrete, additive capabilities which emerge in independent phase transitions. Finally, several works \citep{varma2023explaining, nanda2023progress} observed the phenomenon of grokking, where a model that has already achieved near-perfect performance on the training data transitions to a mechanism that generalizes.

While only a minority of circuits-style analysis has focused on model development, a wealth of results in the field of developmental interpretability have characterized the training dynamics of language models using other units of analysis. Behavioral analysis has shown that many model capabilities, such as predicting parts of speech and individual tokens, are learned during independent phase transitions characterized by varying slopes and points of maximum curvature \citep{chiang-etal-2020-pretrained, wei2022emergent}. \citet{raghu2017svcca} proposed singular vector canonical correlation analysis, a tool which allows different model layers and networks to be compared, and showed that early network layers converge earlier in training than late layers. \citet{saphra2020lstms, saphra-lopez-2020-lstms} found that LSTM models learn hierarchical sequential features in natural language bottom up - shorter sequences are learned first and form the basis for the representation of longer sequences that contain the shorter ones. Finally, \citet{saxe2019semantic} found that the features are learned in an order determined by the magnitude of the corresponding singular value in the model.

\citet{dodge2020finetuning} found that differences in random parameter initializations and training data order contribute comparably to differences between learned representations of fine-tuning runs of the same model on identical data, and \citet{mccoy2020berts} found that identically trained BERT transformer models are consistent with this finding, varying widely in generalization performance. \citet{juneja2023linear} found that models with matching generalization strategies exist in the same basin in parameter space, connected by a simple polygonal chain of two line segments along which loss is non-increasing.

Most relevantly to our work, \citet{class_selective_neurons} found that residual neural networks (of which decoder-only transformers are a special case), when trained on an image classification task, form class-selective neurons, i.e. neurons that systematically fire on instances of an output class. The neurons form early in training and are later unlearned. The importance of the neurons, determined by zero ablation of single neurons, shows the same trend of early growth and then dissipation. The temporary context neurons we observed do not perform class selection but the general pattern of learned and unlearned neurons is similar.

\section{Discussion}
\paragraph{Contributions}

In this article, we presented experiments analyzing the formation and utilization of German context neurons. We found evidence of what we defined as a second-order circuit, a circuit that forms gradually after its constituent circuits are already present, instead of being learned rapidly in a phase transition. The second-order circuit formation does not correspond to a reduction in loss, raising the question of what purpose it serves. Second-order circuits represent a novel insight into the learning behavior of language models and provides a valuable case study that does not seem to fit into the current understanding of circuit formation.

To understand the formation of the German context neuron, we analyzed its direct and indirect effects throughout training. This led to the discovery of the contextual n-gram circuit, a circuit that consists of the German context neuron and neurons in later feed-forward layers, and predicts German n-grams if the context neuron is active. We found that both the context neuron and many German n-grams are learned independently early on in training, but that both the importance of the context neuron and the dependence of specific German n-grams on the context neuron grow gradually throughout training.

Additionally, we identified a surprisingly large cohort of highly accurate context neurons that are learned during a phase transition early on in training. Most of these context neurons are quickly unlearned again. Contrary to our expectations, these neurons had only a very low impact on loss when ablated, raising the question of why they are learned in the first place. This motif is reminiscent of the class-selective neurons \citep{class_selective_neurons} found in image classification models; the primary differences are that our neurons are not class-selective and instead seem to represent the high-level "is German" feature, and a few of the neurons are not unlearned and gradually become important to the model's performance.

We find these results fascinating and unexpected. While we do not claim to fully understand how and why the context neuron is learned, we believe our results offer important glimpses into the training dynamics surrounding feed-forward layer circuits. 
The slow and gradual formation of the contextual n-gram circuit adds nuance to the quantization model of neural scaling \citep{michaud2023quantization}, which postulates that every model capability can be broken down into composite parts which are learned in short phase transitions. 

Our work adds two complexities to this story. First, the universal phase transition in which both the n-gram circuit and the German detection circuit were learned could plausibly be caused by the learning rate warm-up, raising the question whether we observed an idiosyncrasy of the optimizer or a natural phase transition. Second, the contextual n-gram circuit did not seem to reduce the model's loss when it formed. While we do not know why the contextual n-gram circuit forms, we suspect it is associated with internal memory management and reducing superposition. It is plausible that circuits that rearrange internal mechanisms do not correspond to phase transitions and are qualitatively different from those which lead to a direct drop in loss. Therefore, while we refrain from claiming that our work falsifies the quantization model, we believe it adds complexity to the picture.

\paragraph{Limitations and future work}
A limitation of our analysis is that we focused on specifically selected n-grams that strongly rely on the German context neuron. While we are confident that predicting contextual n-grams is an important function of the context neuron, we are uncertain whether it is the only circuit the neuron contributes to. Additionally, our analysis could be extended to other context neurons, model sizes, and models outside the Pythia series.

One exciting direction for future work is to fully reverse-engineer the role of context neurons in language models, either by comprehensively identifying all circuits the neurons are involved in or by proving that unigram and n-gram effects are sufficient to explain their effects. 

Another direction for future work is to investigate further the emergence and role of the early context neurons. We do not yet understand why context neurons form early even though they do not seem to be utilized well at that point, nor do we understand why there are so many early context neurons that are quickly repurposed.

\section*{Statements and Declarations}
\paragraph{Acknowledgements}

This work was supported by funding and mentorship of the Summer 2023 SERI MATS program.

The authors thank the developers of TransformerLens \citep{nandatransformerlens2022}, an open-source library for mechanistic interpretability.

The authors thank Naomi Saphra for helpful discussions around related work, and Oskar Hollinsworth, Evan Hockings, Jett Janiak, and James Dao for their reviews.

\paragraph{Author contributions}
Lovis Heindrich and Lucia Quirke led the project, implemented all experiments and wrote the paper. Neel Nanda was the main supervisor for this project, Wes Gurnee co-supervised. Both Neel and Wes provided feedback and guidance throughout. 

\paragraph{Data availability}
Our data, experiment code, and code to reproduce our visualizations is available on GitHub: \url{https://github.com/luciaquirke/contextual-ngrams}.

\medskip{
\small

\bibliographystyle{apalike}
\bibliography{neurips_2023}

\begin{thebibliography}{}

\bibitem[Biderman et~al., 2023]{pythia}
Biderman, S., Schoelkopf, H., Anthony, Q., Bradley, H., O'Brien, K., Hallahan, E., Khan, M.~A., Purohit, S., Prashanth, U.~S., Raff, E., Skowron, A., Sutawika, L., and van~der Wal, O. (2023).
\newblock Pythia: A suite for analyzing large language models across training and scaling.

\bibitem[Bills et~al., 2023]{bills2023language}
Bills, S., Cammarata, N., Mossing, D., Tillman, H., Gao, L., Goh, G., Sutskever, I., Leike, J., Wu, J., and Saunders, W. (2023).
\newblock Language models can explain neurons in language models.
\newblock \url{https://openaipublic.blob.core.windows.net/neuron-explainer/paper/index.html}.

\bibitem[Bricken et~al., 2023]{bricken2023monosemanticity}
Bricken, T., Templeton, A., Batson, J., Chen, B., Jermyn, A., Conerly, T., Turner, N., Anil, C., Denison, C., Askell, A., Lasenby, R., Wu, Y., Kravec, S., Schiefer, N., Maxwell, T., Joseph, N., Hatfield-Dodds, Z., Tamkin, A., Nguyen, K., McLean, B., Burke, J.~E., Hume, T., Carter, S., Henighan, T., and Olah, C. (2023).
\newblock Towards monosemanticity: Decomposing language models with dictionary learning.
\newblock {\em Transformer Circuits Thread}.
\newblock https://transformer-circuits.pub/2023/monosemantic-features/index.html.

\bibitem[Burns et~al., 2022]{burns2022discovering}
Burns, C., Ye, H., Klein, D., and Steinhardt, J. (2022).
\newblock Discovering latent knowledge in language models without supervision.
\newblock {\em arXiv preprint arXiv:2212.03827}.

\bibitem[Cammarata et~al., 2020]{cammarata2020thread:}
Cammarata, N., Carter, S., Goh, G., Olah, C., Petrov, M., Schubert, L., Voss, C., Egan, B., and Lim, S.~K. (2020).
\newblock Thread: Circuits.
\newblock {\em Distill}.
\newblock https://distill.pub/2020/circuits.

\bibitem[Chan et~al., 2022]{chan2022causal}
Chan, L., Garriga-Alonso, A., Goldwosky-Dill, N., Greenblatt, R., Nitishinskaya, J., Radhakrishnan, A., Shlegeris, B., and Thomas, N. (2022).
\newblock Causal scrubbing, a method for rigorously testing interpretability hypotheses.
\newblock {\em AI Alignment Forum}.
\newblock \url{https://www.alignmentforum.org/posts/JvZhhzycHu2Yd57RN/causal-scrubbing-a-method-for-rigorously-testing}.

\bibitem[Chiang et~al., 2020]{chiang-etal-2020-pretrained}
Chiang, C.-H., Huang, S.-F., and Lee, H.-y. (2020).
\newblock {P}retrained language model embryology: {T}he birth of {ALBERT}.
\newblock In {\em Proceedings of the 2020 Conference on Empirical Methods in Natural Language Processing (EMNLP)}, pages 6813--6828, Online. Association for Computational Linguistics.

\bibitem[Chicco1 and Jurman, 2023]{why_mcc}
Chicco1, D. and Jurman, G. (2023).
\newblock The advantages of the matthews correlation coefficient (mcc) over f1 score and accuracy in binary classification evaluation.
\newblock {\em BMC Genomics}.

\bibitem[Dodge et~al., 2020]{dodge2020finetuning}
Dodge, J., Ilharco, G., Schwartz, R., Farhadi, A., Hajishirzi, H., and Smith, N. (2020).
\newblock Fine-tuning pretrained language models: Weight initializations, data orders, and early stopping.

\bibitem[Elhage et~al., 2022a]{elhage2022solu}
Elhage, N., Hume, T., Olsson, C., Nanda, N., Henighan, T., Johnston, S., ElShowk, S., Joseph, N., DasSarma, N., Mann, B., Hernandez, D., Askell, A., Ndousse, K., Jones, A., Drain, D., Chen, A., Bai, Y., Ganguli, D., Lovitt, L., Hatfield-Dodds, Z., Kernion, J., Conerly, T., Kravec, S., Fort, S., Kadavath, S., Jacobson, J., Tran-Johnson, E., Kaplan, J., Clark, J., Brown, T., McCandlish, S., Amodei, D., and Olah, C. (2022a).
\newblock Softmax linear units.
\newblock {\em Transformer Circuits Thread}.
\newblock https://transformer-circuits.pub/2022/solu/index.html.

\bibitem[Elhage et~al., 2022b]{toy_models}
Elhage, N., Hume, T., Olsson, C., Schiefer, N., Henighan, T., Kravec, S., Hatfield-Dodds, Z., Lasenby, R., Drain, D., Chen, C., Grosse, R., McCandlish, S., Kaplan, J., Amodei, D., Wattenberg, M., and Olah, C. (2022b).
\newblock Toy models of superposition.

\bibitem[Elhage et~al., 2021]{elhage2021mathematical}
Elhage, N., Nanda, N., Olsson, C., Henighan, T., Joseph, N., Mann, B., Askell, A., Bai, Y., Chen, A., Conerly, T., DasSarma, N., Drain, D., Ganguli, D., Hatfield-Dodds, Z., Hernandez, D., Jones, A., Kernion, J., Lovitt, L., Ndousse, K., Amodei, D., Brown, T., Clark, J., Kaplan, J., McCandlish, S., and Olah, C. (2021).
\newblock A mathematical framework for transformer circuits.
\newblock {\em Transformer Circuits Thread}.
\newblock https://transformer-circuits.pub/2021/framework/index.html.

\bibitem[Gao et~al., 2020]{pile}
Gao, L., Biderman, S., Black, S., Golding, L., Hoppe, T., Foster, C., Phang, J., He, H., Thite, A., Nabeshima, N., Presser, S., and Leahy, C. (2020).
\newblock The pile: An 800gb dataset of diverse text for language modeling.

\bibitem[Geva et~al., 2022]{geva2022transformer}
Geva, M., Caciularu, A., Wang, K.~R., and Goldberg, Y. (2022).
\newblock Transformer feed-forward layers build predictions by promoting concepts in the vocabulary space.

\bibitem[Geva et~al., 2021]{geva-etal-2021-transformer}
Geva, M., Schuster, R., Berant, J., and Levy, O. (2021).
\newblock Transformer feed-forward layers are key-value memories.
\newblock In {\em Proceedings of the 2021 Conference on Empirical Methods in Natural Language Processing}, pages 5484--5495, Online and Punta Cana, Dominican Republic. Association for Computational Linguistics.

\bibitem[Gurnee et~al., 2023]{gurnee2023finding}
Gurnee, W., Nanda, N., Pauly, M., Harvey, K., Troitskii, D., and Bertsimas, D. (2023).
\newblock Finding neurons in a haystack: Case studies with sparse probing.
\newblock {\em arXiv preprint arXiv:2305.01610}.

\bibitem[Juneja et~al., 2023]{juneja2023linear}
Juneja, J., Bansal, R., Cho, K., Sedoc, J., and Saphra, N. (2023).
\newblock Linear connectivity reveals generalization strategies.

\bibitem[Leavitt and Morcos, 2020]{leavitt2020falsifiable}
Leavitt, M.~L. and Morcos, A. (2020).
\newblock Towards falsifiable interpretability research.

\bibitem[McCoy et~al., 2020]{mccoy2020berts}
McCoy, R.~T., Min, J., and Linzen, T. (2020).
\newblock Berts of a feather do not generalize together: Large variability in generalization across models with similar test set performance.

\bibitem[McGrath et~al., 2023]{mcgrath2023hydra}
McGrath, T., Rahtz, M., Kramar, J., Mikulik, V., and Legg, S. (2023).
\newblock The hydra effect: Emergent self-repair in language model computations.

\bibitem[Meng et~al., 2023]{meng2023locating}
Meng, K., Bau, D., Andonian, A., and Belinkov, Y. (2023).
\newblock Locating and editing factual associations in gpt.

\bibitem[Michaud et~al., 2023]{michaud2023quantization}
Michaud, E.~J., Liu, Z., Girit, U., and Tegmark, M. (2023).
\newblock The quantization model of neural scaling.

\bibitem[Nanda and Bloom, 2022]{nandatransformerlens2022}
Nanda, N. and Bloom, J. (2022).
\newblock Transformerlens.

\bibitem[Nanda et~al., 2023]{nanda2023progress}
Nanda, N., Chan, L., Lieberum, T., Smith, J., and Steinhardt, J. (2023).
\newblock Progress measures for grokking via mechanistic interpretability.

\bibitem[nostalgebraist, 2020]{nostalgebraist2020}
nostalgebraist (2020).
\newblock interpreting gpt: the logit lens.
\newblock \url{https://www.lesswrong.com/posts/AcKRB8wDpdaN6v6ru/interpreting-gpt-the-logit-lens}.
\newblock Accessed: 28/09/2023.

\bibitem[Olah, 2022]{olah2022mechanistic}
Olah, C. (2022).
\newblock Mechanistic interpretability, variables, and the importance of interpretable bases. transformer circuits thread (june 27).

\bibitem[Olsson et~al., 2022]{olsson2022incontext}
Olsson, C., Elhage, N., Nanda, N., Joseph, N., DasSarma, N., Henighan, T., Mann, B., Askell, A., Bai, Y., Chen, A., Conerly, T., Drain, D., Ganguli, D., Hatfield-Dodds, Z., Hernandez, D., Johnston, S., Jones, A., Kernion, J., Lovitt, L., Ndousse, K., Amodei, D., Brown, T., Clark, J., Kaplan, J., McCandlish, S., and Olah, C. (2022).
\newblock In-context learning and induction heads.

\bibitem[Pearl, 2001]{pearldirect2001}
Pearl, J. (2001).
\newblock Direct and indirect effects.
\newblock In {\em Proceedings of the Seventeenth Conference on Uncertainty in Artificial Intelligence}, UAI'01, page 411–420, San Francisco, CA, USA. Morgan Kaufmann Publishers Inc.

\bibitem[Raghu et~al., 2017]{raghu2017svcca}
Raghu, M., Gilmer, J., Yosinski, J., and Sohl-Dickstein, J. (2017).
\newblock Svcca: Singular vector canonical correlation analysis for deep learning dynamics and interpretability.

\bibitem[Ranadive et~al., 2023]{class_selective_neurons}
Ranadive, O., Thakurdesai, N., Morcos, A.~S., and Matthew~Leavitt, S.~D. (2023).
\newblock On the special role of class-selective neurons in early training.
\newblock {\em Transactions on Machine Learning Research}.

\bibitem[Räuker et~al., 2023]{räuker2023transparent}
Räuker, T., Ho, A., Casper, S., and Hadfield-Menell, D. (2023).
\newblock Toward transparent ai: A survey on interpreting the inner structures of deep neural networks.

\bibitem[Saphra and Lopez, 2020a]{saphra-lopez-2020-lstms}
Saphra, N. and Lopez, A. (2020a).
\newblock {LSTM}s compose{---}and {L}earn{---}{B}ottom-up.
\newblock In {\em Findings of the Association for Computational Linguistics: EMNLP 2020}, pages 2797--2809, Online. Association for Computational Linguistics.

\bibitem[Saphra and Lopez, 2020b]{saphra2020lstms}
Saphra, N. and Lopez, A. (2020b).
\newblock Lstms compose (and learn) bottom-up.

\bibitem[Saxe et~al., 2019]{saxe2019semantic}
Saxe, A.~M., McClelland, J.~L., and Ganguli, S. (2019).
\newblock A mathematical theory of semantic development in deep neural networks.

\bibitem[Taha and Hanbury, 2015]{taha2015metrics}
Taha, A.~A. and Hanbury, A. (2015).
\newblock Metrics for evaluating 3d medical image segmentation: analysis, selection, and tool.
\newblock {\em BMC Med Imaging}.

\bibitem[Varma et~al., 2023]{varma2023explaining}
Varma, V., Shah, R., Kenton, Z., Kramár, J., and Kumar, R. (2023).
\newblock Explaining grokking through circuit efficiency.

\bibitem[Vig et~al., 2020]{vig2020investigating}
Vig, J., Gehrmann, S., Belinkov, Y., Qian, S., Nevo, D., Singer, Y., and Shieber, S. (2020).
\newblock Investigating gender bias in language models using causal mediation analysis.
\newblock {\em Advances in neural information processing systems}, 33:12388--12401.

\bibitem[Wang et~al., 2022]{wang2022interpretability}
Wang, K., Variengien, A., Conmy, A., Shlegeris, B., and Steinhardt, J. (2022).
\newblock Interpretability in the wild: a circuit for indirect object identification in gpt-2 small.

\bibitem[Wei et~al., 2022]{wei2022emergent}
Wei, J., Tay, Y., Bommasani, R., Raffel, C., Zoph, B., Borgeaud, S., Yogatama, D., Bosma, M., Zhou, D., Metzler, D., Chi, E.~H., Hashimoto, T., Vinyals, O., Liang, P., Dean, J., and Fedus, W. (2022).
\newblock Emergent abilities of large language models.

\end{thebibliography}
}

\end{document}